\documentclass[10pt,twocolumn,letterpaper]{article}

\usepackage[pagenumbers]{wacv} 

\usepackage{graphicx}
\usepackage{amsmath}
\usepackage{amssymb}
\usepackage{booktabs}

\usepackage{algorithm2e}
\usepackage{stmaryrd}

\usepackage[pagebackref,breaklinks,colorlinks]{hyperref}

\usepackage[capitalize]{cleveref}
\crefname{section}{Sec.}{Secs.}
\Crefname{section}{Section}{Sections}
\Crefname{table}{Table}{Tables}
\crefname{table}{Tab.}{Tabs.}

\def\wacvPaperID{511} 
\def\confName{WACV}
\def\confYear{2024}

\begin{document}

\title{Gradual Source Domain Expansion for Unsupervised Domain Adaptation}

\author{Thomas Westfechtel$^1$, Hao-Wei Yeh$^1$, Dexuan Zhang$^1$, Tatsuya Harada$^{1,2}$ \\
$^1$The University of Tokyo \hspace*{1cm} $^2$RIKEN\\
Tokyo, Japan\\
{\tt\small \{thomas,yeh,dexuan.zhang,harada\}@mi.t.u-tokyo.ac.jp}}

\maketitle

\begin{abstract}
Unsupervised domain adaptation (UDA) tries to overcome the need for a large labeled dataset by transferring knowledge from a source dataset, with lots of labeled data, to a target dataset, that has no labeled data. Since there are no labels in the target domain, early misalignment might propagate into the later stages and lead to an error build-up.
In order to overcome this problem, we propose a gradual source domain expansion (GSDE) algorithm. GSDE trains the UDA task several times from scratch, each time reinitializing the network weights, but each time expands the source dataset with target data. In particular, the highest-scoring target data of the previous run are employed as pseudo-source samples with their respective pseudo-label. Using this strategy, the pseudo-source samples induce knowledge extracted from the previous run directly from the start of the new training. This helps align the two domains better, especially in the early training epochs.
In this study, we first introduce a strong baseline network and apply our GSDE strategy to it. We conduct experiments and ablation studies on three benchmarks (Office-31, OfficeHome, and DomainNet) and outperform state-of-the-art methods. We further show that the proposed GSDE strategy can improve the accuracy of a variety of different state-of-the-art UDA approaches.
\end{abstract}

\section{Introduction}

Deep neural networks have advanced most computer vision tasks greatly. However, large labeled datasets are required to train these networks. While there is a variety of large datasets available online, most times there exists a domain shift between the available data and the target data, for which the network will be employed. This can be overcome by labeling the target data and finetuning the network on it, but the labeling process is very tedious and costly. Unsupervised domain adaptation (UDA) overcomes the need to label the target data by transferring knowledge from a labeled source dataset to an unlabeled target dataset. 

One problem of UDA is early alignment error build-up. At the start of the training, the classification network is neither aligned to the source nor the target domain. Usually, warm-up training using only the source data or a progressive learning rate for the adaptation task is employed. However, this mostly ignores the target data, meaning that the classifier only aligns to the source data. In this work, we introduce a strong prior in the form of pseudo-source data that is instilled right from the beginning of the training process. In particular, we start the adaptation process $N$ times, each time reinitializing the network weights. Each time the training process restarts, we use the most confident predictions with their pseudo-labels of the previous run (i.e. the class predictions) and introduce them into the source dataset as pseudo-source data. This mainly gives two advantages. Firstly, the pseudo-source data allow the classifier to align to the target data from the beginning of the training, and secondly, the pseudo-source data act as guidance for the target data during the domain alignment process.

While it is beneficial to have a strong prior, meaning a large amount of target data employed as pseudo-source data, this also increases the chance of misclassified data being employed as pseudo-source, which would be harmful to the adaptation. To mitigate this dilemma, the GSDE algorithm is run iteratively, each time increasing the amount of target data used as pseudo-source data.
For the $n$-th run we introduce the $\frac{n-1}{N}$ highest scoring target data into the source dataset as pseudo source data. Therefore, with each run we introduce a stronger prior to the training, letting the network early on align to both source and target data.

In this paper, we first introduce a strong baseline network that consists of four different losses: classification loss of source (and pseudo-source) data, domain adversarial loss, semantic loss, a semi-supervised loss. We then apply the proposed Gradual Source Domain Expansion (GSDE) strategy to it. In addition, we present additional improvements, in particular the use of multiple bottlenecks and an advanced scoring technique for pseudo-labels. Finally, we also show that the GSDE strategy works well with other state-of-the-art domain adaptation methods, based on a variety of adaption principles.

Our main contributions are:
\begin{itemize}
\item We introduce a strong baseline network consisting of four different losses.
\item We introduce a gradual source domain expansion strategy (GSDE), that allows the network to overcome the problem of early misalignment build-up by introducing a strong prior based on previous runs. Each time the network weights are reinitialized to counter the early alignment error build-up.
\item We show the effectiveness of our algorithm on three datasets (Office-31, Office-Home, and DomainNet) and further evaluate the method in various ablation studies.
\item We validate that our GSDE strategy also works with a variety of UDA methods, that are based on different adaptation strategies, and show that the addition of GSDE yields significant increases in accuracy.
\end{itemize}

 \begin{figure}[t]
   \centering
        \begin{subfigure}[htb]{0.225\textwidth}
                \includegraphics[width=\textwidth ]{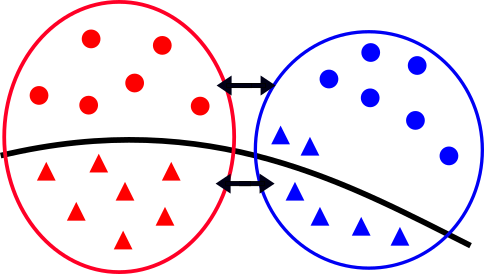}
                \caption{Start of alignment.}
                \label{fig:ad-start}       
        \end{subfigure}
        \hfill
        \begin{subfigure}[htb]{0.225\textwidth}
                \includegraphics[width=\textwidth ]{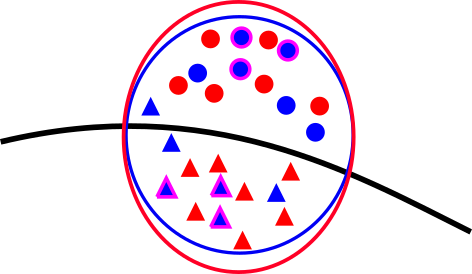}
                \caption{Finish of alignment.}
                \label{fig:ad-end}
        \end{subfigure}   
        \hfill
              \begin{subfigure}[htb]{0.225\textwidth}
                \includegraphics[width=\textwidth ]{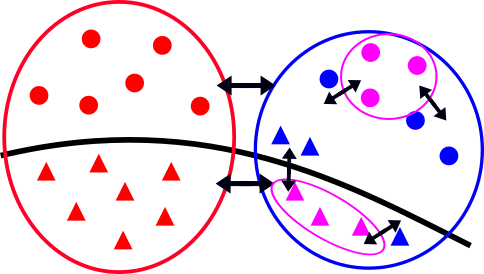}
                \caption{Start of alignment with pseudo-source.}
                \label{fig:ad-start-2}       
        \end{subfigure}
        \hfill
        \begin{subfigure}[htb]{0.225\textwidth}
                \includegraphics[width=\textwidth ]{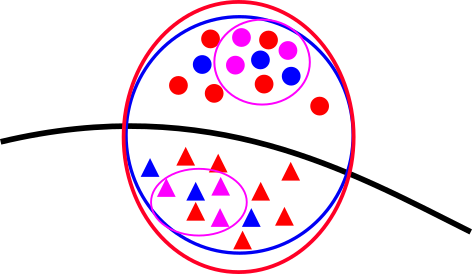}
                \caption{Finish of alignment with pseudo-source.}
                \label{fig:ad-end-2}
        \end{subfigure}   \\
        \caption{Concept of our idea for adversarial adaptation. During the alignment of the source and target domain (a), some samples might end up misclassified (b). Using the confident target samples as pseudo-source samples in a new run (c) helps to guide the adaptation, leading to a better adaptation (d).} 
        \label{fig:Concept-Ad}
  \end{figure}   

\section{Related Work}

One main strategy to solve the problem of unsupervised domain adaptation (UDA) is to align the feature representations of source and target domain. Surveys for this task can be found in \cite{s-1}, \cite{s-2}. Probably the most common strategy to achieve this is by using an adversarial approach. Usually, a domain classifier is employed to distinguish whether the feature space of an image belongs to the source or target domain. Domain-adversarial neural network (DANN) \cite{DANN} introduced a gradient reversal layer before the domain classifier, so that the feature extractor is trained to extract features that are indistinguishable for the domain classifier. Conditional domain adversarial networks (CDAN) \cite{CDAN} extends this method by multilinear conditioning the domain classifier with the classifier predictions. A lot of researchers have built up on DANN or CDAN. \\
\cite{ASAN} introduces a spectral adaptation to CDAN. \cite{GradSync} adds group- and class-wise domain classifiers to DANN and synchronizes the gradient between the different domain classifiers. 
Moving semantic transfer network \cite{MSTN} extends DANN with a moving semantic loss. The method creates class representations for both domains and each class in the feature space, which are updated with each sample during the training process. The distance between the source and target feature representation of a class is used as domain adaptation loss. 

   \begin{figure}[t]
   \centering      
   \vspace*{0.3cm}
                \begin{subfigure}[htb]{0.225\textwidth}
                \includegraphics[width=\textwidth ]{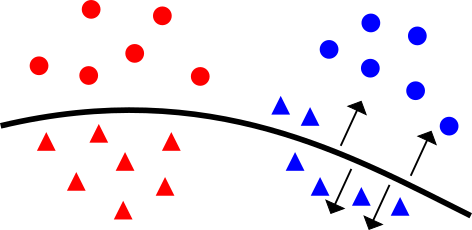}
                \caption{Start of training.}
                \label{fig:em-start}       
        \end{subfigure}
        \hfill
        \begin{subfigure}[htb]{0.225\textwidth}
                \includegraphics[width=\textwidth ]{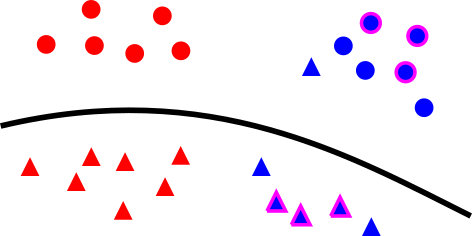}
                \caption{Finish of training.}
                \label{fig:em-end}
        \end{subfigure}   \\
        \vspace*{0.15cm}
              \begin{subfigure}[htb]{0.225\textwidth}
                \includegraphics[width=\textwidth ]{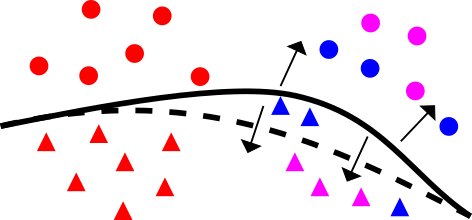}
                \caption{Start of training with pseudo-source.}
                \label{fig:em-start-2}       
        \end{subfigure}
        \hfill
        \begin{subfigure}[htb]{0.225\textwidth}
        \vspace*{0.3cm}
                \includegraphics[width=\textwidth ]{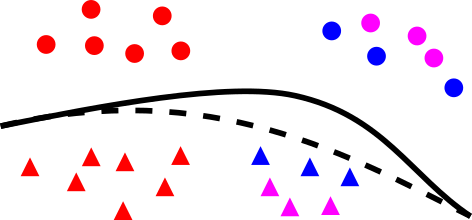}
                \caption{Finish of training with pseudo-source.}
                \label{fig:em-end-2}  
         \end{subfigure}     \\
                         \begin{subfigure}[htb]{0.34\textwidth}
                \includegraphics[width=\textwidth ]{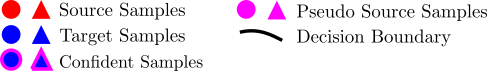}
                \caption{Legend.}
                \label{fig:leg}
        \end{subfigure}     
        \caption{Concept of our idea for adaptation using entropy minimization. As the classifier is not aligned to the target samples (a), samples might end up being misclassified (b). Using the confident target samples as pseudo-source samples allows the classifier (and its decision boundary) to adjust to the target samples (c), resulting in a better adaptation (d).} 
        \label{fig:Concept-EM}
  \end{figure}

Another approach for UDA is information maximization or entropy minimization. \cite{SHOT} exploits both information maximization and self-supervised pseudo-labeling to implicitly align the representations of both domains. \cite{SENTRY} employs data augmentation and minimizes the entropy if the predictions of the different data augmentations are consistent, or maximizes the entropy otherwise. 

One problem of UDA is early alignment error build-up. At the start of the training, the classification network is neither aligned to the source nor the target domain. Adopting target to source in this state could introduce a misalignment that propagates to the later stages of the training. Adversarial methods usually employ a progressive learning rate \cite{DANN}, \cite{CDAN}, \cite{MSTN}, other methods often employ warm-up training using only the source data \cite{SENTRY}, source-free UDA methods even decouple the training of the model with source data with the adaptation phase using only target data \cite{SHOT}, \cite{George}, while still achieving state-of-the-art results.

However, we argue, that this introduces a strong bias towards the source data as the network mostly or only relies on source data during the early stages of the training, resulting in a network that is not well aligned to the target data, which can lead to an error-build up.

To overcome this problem \cite{PFAN} introduces a progressive feature alignment network. In particular employing an easy-to-hard transfer strategy, that adapts easy, or well-aligned, samples first and progressively introduces harder samples. \cite{CAN} employed a clustering method in combination with a class-aware sampler that excludes hard (ambiguous) samples and classes from the training during the early stages. 

In contrast to this, we introduce pseudo-source data into the source dataset as a strong prior. This allows to draw on the knowledge extracted from previous runs to adapt to the target data even early on in the training.

\section{Methodology}

In unsupervised domain adaptation, the task is to mitigate the domain shift between a source and target domain. For the source domain $\mathcal D_s$ a set of $n_s$ labeled samples $\mathcal D_s = {(x_{i,s},y_{i,s})}_{i=1}^{n_s}$ is given, where $x_{i,s}$ donates a sample with the corresponding label $y_{i,s}$. For the target domain $\mathcal D_t$ only the samples are given without any labels $\mathcal D_t = {(x_{i,t})}_{i=1}^{n_t}$. The goal is to estimate the labels for the target domain $\hat y_{i,t}$ by exploiting the shared feature space that is similar, but different. In our work, we tackle the vanilla or closed-set setting, where the source and target domain have identical label classes $\mathcal C_s = \mathcal C_t$.

\begin{algorithm}[t]
\caption{Algorithm of the proposed gradual source domain expansion.}
\SetKwInOut{Input}{Input}
\Input{Source and target dataset: $\mathcal D_s, \mathcal D_t$}
\tcc{Iteratively train network from scratch for $N$ runs} 
\For{$n=1; n\leq N$}{
\tcc{Create expanded source dataset} 
 $D'_t = D_t \in \mathrm{top}  \frac{n-1}{N}$ of $\hat p(y_T)$ \\
 $D'_s = D_s \cup D'_t$ \\
 \tcc{Train network}
 Initialize dataloader $D'_s$ and $D_t$ \\
 Initialize feature extractor and classifier $G_c$,$G_f$ \\
 Train network $G_c$,$G_f$\\
 \tcc{Calculate scores for next run} 
Calculate predictions for target data \\
$\hat p(y_T) \shortleftarrow G_c(G_f(x_T)) $ \\
}
\label{alg::GSE}
\end{algorithm}

\subsection{Gradual Source Domain Expansion}
The main contribution of this work is the introduction of the gradual source domain expansion (GSDE) strategy. The strategy trains the network several times from scratch, each time reinitializing the network weights, and increasing the amount of pseudo-source samples and therefore increasing the instilled prior knowledge from the previous run. 
In detail, we train the network $N$ times from scratch, each time reinitializing the network weights, and for each consecutive run $n$ we expand the source domain by adding target domain samples to it:  $\mathcal D'(n)_s = D_s \cup D(n)'_t$, where $D(n)'_t$ is a subset from $D_t$. The samples from the subset $D(n)'_t$ are assigned pseudo-labels according to the predictions of the previous run. $\mathcal D'_t = {(x_{i,t},\hat y_{i,t})}_{i=1}^{n'_t}$. 

While it is beneficial to have a strong prior, meaning a large amount of target data employed as pseudo-source data, it also increases the chance of misclassified data being employed as pseudo-source. This in turn would be harmful for the adaptation. In order to mitigate this dilemma, the GSDE algorithm is run iteratively. In the first few runs only a few, high-scoring samples are employed as pseudo-source data, thus decreasing the possibility of employing misclassified samples. In practice, for the $n$-th run, we employ the $\frac{n-1}{N}$ highest scoring target samples with their respective pseudo-labels. The algorithm can be found in Alg. \ref{alg::GSE}. 

\subsection{Motivation and Intuitive Explanation}
For domain adaptation task, adversarial adaptation and entropy minimization are two adaptation strategies that are commonly used. The aim of adversarial adaptation is to generate domain-invariant features, aligning the feature space of source and target domain. Entropy minimization on the other hand tries to minimize the entropy of the target samples, in effect moving the samples away from the decision boundary.

The concept of our idea is displayed in Fig. \ref{fig:Concept-Ad} for adversarial adaptation and Fig. \ref{fig:Concept-EM} for entropy minimization. In the beginning, the feature distribution of the source and target dataset are apart. Through adversarial training, the feature extractor is trained to generate domain invariant features, meaning that source domain and target domain are moved towards each other to encompass the same space in the feature space. However, this process is class-agnostic. This means that while the same feature space is occupied, some samples end up on the wrong side of the classification boundary as shown in Fig. \ref{fig:ad-end}. Using the confident samples from the previous run as pseudo-source samples helps to align the features within the target domain (Fig. \ref{fig:ad-start-2}). Target samples are aligned simultaneously to the pseudo-source and original source data. The pseudo-source data can be seen as guidance or anchor for the alignment process (Fig. \ref{fig:ad-end-2}).

In the case of entropy minimization, one problem is that the classifier might not be well aligned with the target data as can be seen in Fig. \ref{fig:em-start}. This often leads to early alignment error build-up, especially for samples that are close to the decision boundary (Fig. \ref{fig:em-end}). Introducing a strong prior in the form of the pseudo-source labels helps to align the decision boundary for the target data (Fig. \ref{fig:em-start-2}), resulting in a decision boundary that is better aligned to the target data early on. Thus resulting in a better adaptation (Fig. \ref{fig:em-end-2}).

Many adaptation methods also employ a mixture of both adaptation strategies.
The GSDE algorithm is run iteratively, employing only a few high-scoring samples as pseudo-source data first, thus decreasing the possibility of employing misclassified samples. With each run the amount of pseudo-source data is increased.
\begin{table*}
  \caption{Accuracy results on Office-31 dataset. Best results are displayed in bold and the runner-up results are underlined. We display the results of the network with and without our proposed Gradual Source Domain Expansion strategy.}
  \label{tab:office31}
  \centering
  \begin{tabular}{llllllll}
    \toprule
    Method & A$\shortrightarrow$W  & D$\shortrightarrow$W & W$\shortrightarrow$D & A$\shortrightarrow$D & D$\shortrightarrow$A  & W$\shortrightarrow$A & Avg \\
    \midrule
    ResNet-50 \cite{ResNet} & 68.4 & 96.7 & 99.3 & 68.9 & 62.5 & 60.7 & 76.1 \\
    \midrule
	MCD \cite{MCD} & 88.6 & 98.5 & \textbf{100.} & 92.2 & 69.5 & 69.7 & 86.5 \\
	MSTN \cite{MSTN} & 91.3 & 98.9 & \textbf{100.} & 90.4 & 72.7 & 65.6 & 86.5 \\
	CDAN+E \cite{CDAN} & 94.1 & 98.6 & \textbf{100.} & 92.9 & 71.0 & 69.3 & 87.7 \\
	SymNets \cite{SymNets} & 90.8 & 98.8 & \textbf{100.} & 93.9 & 74.6 & 72.5 & 88.4 \\
 MJE \cite{ZHA} & 91.9 & 99.0 & \textbf{100.} & 93.7 & 76.1 & 77.8 & 89.8 \\
	BIWAA-I \cite{BIWA} & 95.6 & 99.0 & \textbf{100.} & 95.4 & 75.9 & 77.3 & 90.5 \\
	CAN  \cite{CAN} & 94.5 & 99.1 & \underline{99.8} & 95.0 & 78.0 & 77.0 & 90.6 \\
	SRDC \cite{SRDC} & 95.7 & \underline{99.2} & \textbf{100.} & \underline{95.8} & 76.7 & 77.1 & 90.8 \\
	FixBi \cite{FixBi} & \underline{96.1} & \textbf{99.3} & \textbf{100.} & 95.0 & \textbf{78.7} & \textbf{79.4} & \underline{91.4} \\
	\midrule
	\textbf{Ours} & 95.8 & \underline{99.2} & \textbf{100.} & 95.6 & 76.0 & 77.2 & 90.6
 \\
	\textbf{Ours+GSDE} & \textbf{96.9} & 98.8 & \textbf{100.} & \textbf{96.7} & \underline{78.3} & \underline{79.2} & \textbf{91.7}
\\

    \bottomrule
  \end{tabular}
  \end{table*}

\section{Adaptation Network}
In this section, we introduce our baseline network which consists of four losses: classification loss $L_C$, adversarial loss $L_{AD}$, semantic loss $L_{MS}$, and semi-self-supervised loss $L_{SS}$.
\begin{equation}
	L = L_C + L_{AD} + L_{MS} + L_{SS}
\end{equation}
The classification loss is the cross-entropy loss and is used for the extended source domain $D'_s$. It should be noted that the other losses also employ the extended source domain as source data.
We chose the three adaptation losses as they complement each other well. The adversarial loss advocates domain invariant features, the semantic loss creates compact representations within each class and increases the distance between representations of different classes, and finally the semi-self-supervised loss promotes augmentation invariant features.

Furthermore, we introduce a multiple bottleneck architecture and an advanced scoring technique for the pseudo-labels as additional improvements.

\subsection{Loss Functions}
\subsubsection{Adversarial loss: $L_{AD}$}

The first adaptation loss of our method is an adversarial loss. We employ the CDAN \cite{CDAN} network for it:
\begin{equation}
L_{AD} = l_{am} \cdot L_{BCE}(G_d((f_i \otimes p_i), d_i))
\end{equation}
where $G_d$ is the domain classification network. $f_i$ are the features of sample $i$, $p_i$ the class probabilities and $d_i$ the domain label. Same as for CDAN, we employ a progressive learning rate $l_{am}$ for the adversarial loss. A gradient reversal layer is employed before the domain classification network in order to invert the training objective, from discriminating the domains to creating indistinguishable domain features.

\subsubsection{Semantic loss: $L_{MS}$}
For the second adaptation loss of our method, we use a moving semantic transfer loss $L_{MS}$. This loss is based on MSTN \cite{MSTN}: 
\begin{equation}
	L_{MSTN} = \sum_{k=1}^K \Phi(C_s^k,C_t^k)
\end{equation}
where $C_s^k$ and $C_t^k$ are the moving centroids of the classes in feature space for source and target data respectively. $\Phi$ is a distance measure. $L_{MSTN}$ aligns the class representations of source and target data within the feature space. Inspired by current deep-clustering-based methods \cite{CAN} \cite{SRDC} we extend the loss to also enlarge the distance between centroids of different classes:
\begin{align}
	L_{MS} = \sum_{k=1}^K l_{am} \cdot \Theta(C_s^k,C_t^k) + \nonumber \\
	\sum_{k=1}^K \sum_{j\neq k}^K \Theta(C_s^k,C_s^j) + l_{am} \cdot \Theta(C_s^k,C_t^j) + l_{am} \cdot \Theta(C_t^k,C_t^j)
\end{align}
The cosine similarity between the centroids is used as function $\Theta$. $l_{am}$ is the progressive learning rate, the same as for the other adaptation losses. Note that $\Theta(C_s^k,C_s^j)$ does not employ $l_{am}$ as it only relies on source data.

\subsubsection{Semi-self-supervised loss: $L_{SS}$}
For the semi-self-supervised loss, we chose the MixMatch algorithm \cite{MixMatch} as it combines consistency regularization, MixUp regularization, and entropy minimization.

\subsection{Additional Improvements}
We further introduce multiple bottlenecks and an advanced scoring of the pseudo labels as additional improvements. 

\subsubsection{Multiple bottlenecks}
Inspired by \cite{HDA} and \cite{GVB}, we employ a multiple bottleneck strategy. Our implementation employs $k$ bottleneck layers in parallel and averages over the output of all bottleneck layers. The use of multiple bottleneck layers that are all initialized differently prevents the bottleneck from converging into a local minima. The output of the multiple bottlenecks is calculated as:
\begin{align}
B(y_{bb}) = \frac{1}{k} \sum_k B_k(y_{bb})
\end{align}
where $y_{bb}$ is the output of the backbone, and $B_k$ is the $k$-th bottleneck.

\subsubsection{Scoring of pseudo labels}
Apart from the probability score of the classifier, we further employ a neighborhood aggregation score and a score based on label propagation. 
\begin{align}
p_{i}^{all} = \frac{1}{3} (p_i + p_{i}^{NA} + p_{i}^{LP})
\end{align}
The neighborhood aggregation score is motivated by \cite{AUX} and finds the $m$ closest target data in the feature space and aggregates their respective classification probability scores. 
\begin{align}
p_{i}^{NA} = \frac{1}{m} \sum_m p(k)
\end{align}
The label propagation score function is based on \cite{LP} and follows the same implementation as \cite{GDA} which is achieved by minimizing the objective:
\begin{align}
\sum_{i=1}^{n} || p_{i}^{LP} - p_i || + \lambda \sum_{i,j}^{n} a_{i,j} ||\frac{p_{i}^{LP}}{\sqrt{d_{ii}}} - \frac{p_{j}^{LP}}{\sqrt{d_{jj}}} ||^2
\end{align}
where $n$ is the amount of both source and target data, $y$ is a one-hot vector with the ground truth label for the source data, and $0$ otherwise, $a_{i,j}$ depicts the cosine similarity between the samples $i$ and $j$.

  \begin{table*}
  \caption{Accuracy results on Office-Home dataset. Best results are displayed in bold and the runner-up results are underlined. We display the results of the network with and without our proposed Gradual Source Domain Expansion strategy.}
  \label{tab:officeHome}
  \centering
  \begin{tabular}{llllllllllllll}
    \toprule
    Method & A$\shortrightarrow$C  & A$\shortrightarrow$P & A$\shortrightarrow$R & 
    C$\shortrightarrow$A & C$\shortrightarrow$P  & C$\shortrightarrow$R &
    P$\shortrightarrow$A & P$\shortrightarrow$C  & P$\shortrightarrow$R &
    R$\shortrightarrow$A & R$\shortrightarrow$C  & R$\shortrightarrow$P &
     Avg \\
    \midrule
    ResNet-50 \cite{ResNet} & 34.9 & 50.0 & 58.0 & 37.4 & 41.9 & 46.2 & 38.5 & 31.2 & 60.4 & 53.9 & 41.2 & 59.9 & 46.1 \\
    \midrule
	MSTN \cite{MSTN} & 49.8 & 70.3 & 76.3 & 60.4 & 68.5 & 69.6 & 61.4 & 48.9 & 75.7 & 70.9 & 55.0 & 81.1 & 65.7 \\
	CDAN+E \cite{CDAN} & 50.7 & 70.6 & 76.0 & 57.6 & 70.0 & 70.0 & 57.4 & 50.9 & 77.3 & 70.9 & 56.7 & 81.6 & 65.8 \\
	GVB-GD \cite{GVB} & 57 & 74.7 & 79.8 & 64.6 & 74.1 & 74.6 & 65.2 & 55.1 & 81.0 & 74.6 & 59.7 & 84.3 & 70.4 \\
	DCAN \cite{DCAN} & 54.5 & 75.7 & \underline{81.2} & 67.4 & 74.0 & 76.3 & 67.4 & 52.7 & 80.6 & 74.1 & 59.1 & 83.5 & 70.5 \\
	BIWAA-I \cite{BIWA} & 56.3 & \underline{78.4} & \underline{81.2} & 68.0 & 74.5 & 75.7 & \underline{67.9} & 56.1 & 81.2 & 75.2 & 60.1 & 83.8 & 71.5 \\
 	SRDC \cite{SRDC} & 52.3 & 76.3 & 81.0 & \underline{69.5} & 76.2 & 78.0 & \textbf{68.7} & 53.8 & 81.7 & \underline{76.3} & 57.1 & 85.0 & 71.3 \\
 MJE \cite{ZHA} & \underline{60.3} & 77.8 & 81.0 & 66.0 & 74.4 & 74.5 & 66.7 & \underline{59.3} & 81.8 & 74.2 & 62.7 & 84.9 & 72.0 \\ 	
 	Sentry \cite{SENTRY} & \textbf{61.8} & 77.4 & 80.1 & 66.3 & 71.6 & 74.7 & 66.8 & \textbf{63.0} & 80.9 & 74.0 & \textbf{66.3} & 84.1 & 72.2 \\
 FixBi \cite{FixBi} & 58.1 & 77.3 & 80.4 & 67.7 & \textbf{79.5} & \underline{78.1} & 65.8 & 57.9 & 81.7 & \textbf{76.4} & \underline{62.9} & \textbf{86.7} & \underline{72.7} \\
 	\midrule
 	 \textbf{Ours} & 54.8 & 76.7 & 80.9 & 68.0 & 76.3 & 77.0 & 66.0 & 55.1 & \underline{81.9} & 75.7 & 59.7 & 83.8 & 71.3 \\
\textbf{Ours+GSDE} & 57.8 & \textbf{80.2} & \textbf{81.9} & \textbf{71.3} & \underline{78.9} & \textbf{80.5} & 67.4 & 57.2 & \textbf{84.0} & 76.1 & 62.5 & \underline{85.7} & \textbf{73.6} \\

    \bottomrule
  \end{tabular}
\end{table*}

\section{Experiments}

We evaluate our proposed method on three different domain adaptation benchmarks, Office-31, Office-Home, and DomainNet. We show that we can improve the baselines significantly. In ablation studies, we further investigate the contribution of the different parts of our proposed algorithm.

\subsection{Setup}
\textbf{Office-31} \cite{O31} is the most popular dataset for real-world domain adaptation. It contains 4,110 images of 31 categories. The domains are Amazon (A), Webcam (W), and DSLR (D). We evaluate all six possible adaptation tasks.

\textbf{Office-Home} \cite{OH} is a more challenging benchmark than Office-31. It contains 15,500 images of 65 categories. The domains are Art (A), Clipart (C), Product (P), and Real-World (R). We evaluate all twelve possible adaptation tasks. 

\textbf{DomainNet} \cite{DN} is a large-scale dataset with about 600,000 images from 6 different domains and 345 different classes. However, as some domains and classes have a considerable amount of mislabeled data, we follow \cite{COAL} and only use a subset of 40 commonly seen classes from the four domains of Real World (R), Clipart (C), Painting (P), and Sketch (S). We evaluate all twelve possible adaptation tasks. Other than for the other two datasets, the target data for adapting and testing are different, furthermore, the per-class accuracy is reported for this dataset. 

\textbf{Implementation details:} 
We built up our implementation on the CDAN implementation of \cite{CDAN}. We use the ResNet-50 \cite{ResNet} architecture as the backbone for all of our experiments. We train each run for $5000$ iterations and employ the final network for the predictions, we do not do any checkpoint selection. We increase the learning rate by a factor of $10$ for all layers that are trained from scratch. We further adopt the learning rate annealing strategy and the progressive discriminator learning strategy $\l_{am}$. We employ $k=5$ bottleneck layers in parallel. The GSDE is executed with a maximum run of $N=5$. 
Each experiment is run for three different seeds.

\subsection{Results}


 \begin{figure}[t]
   \centering
        \begin{subfigure}[htb]{0.2\textwidth}
                \includegraphics[width=\textwidth ]{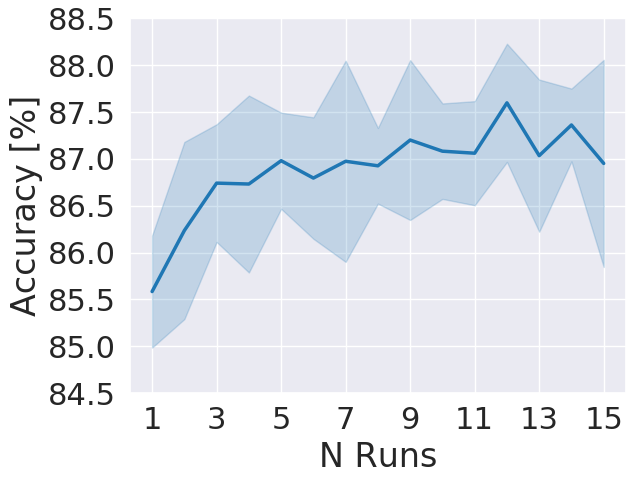}
                
                \caption{Accuracy over max run.}
                \label{fig:maxRun}       
        \end{subfigure}
        \hfill
                \begin{subfigure}[htb]{0.25\textwidth}
                \vspace*{0.13cm}
                \includegraphics[width=\textwidth ]{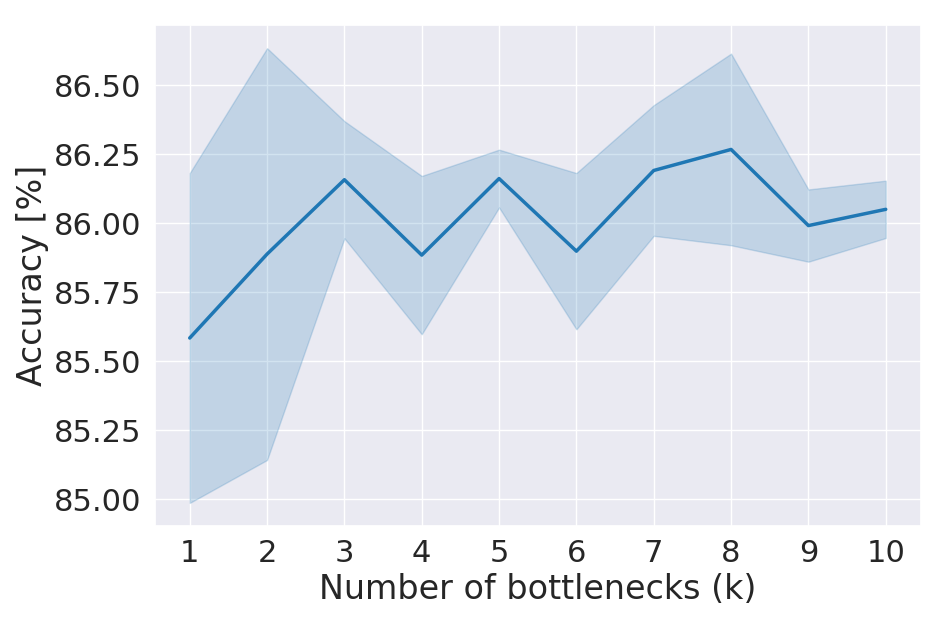}
                \caption{Accuracy for different number of bottlenecks.}
                \label{fig:bneck}
        \end{subfigure}    \\
        \begin{subfigure}[htb]{0.35\textwidth}
                \includegraphics[width=\textwidth ]{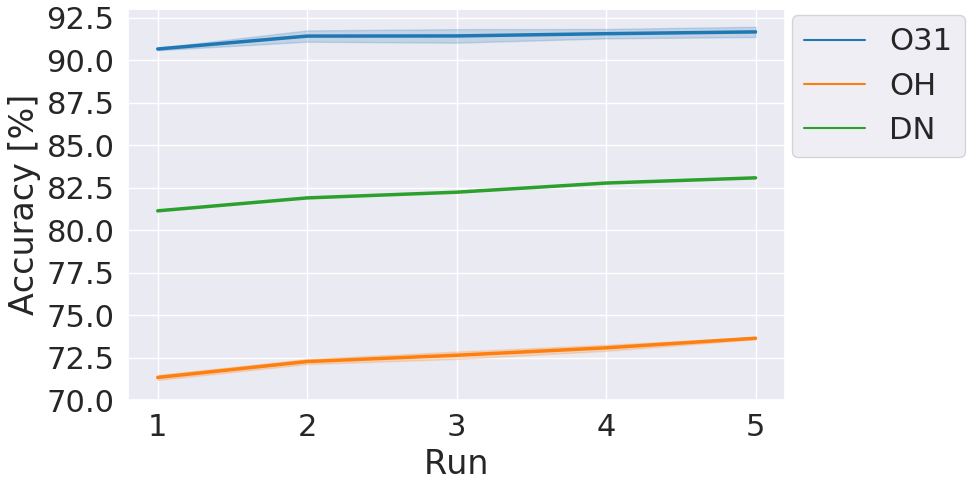}
                \caption{Accuracy over runs.}
                \label{fig:accRun}
        \end{subfigure}

        \caption{Accuracy for different max runs on a subset of the Office31 dataset (left). Accuracy for different numbers of bottlenecks (bottom) over the same subset. Average accuracy for the datasets for max run of $N=5$ for the consecutive runs $n$. (right). } 
        \label{fig:Run}
  \end{figure}

\textbf{Results for Office-31:} The results for the Office-31 dataset are shown in Tab. \ref{tab:office31}. Our base network already performs quite well, only being outperformed by SRDC and FixBi. Using the GSDE strategy proposed in this paper, the accuracy increases by $1.1\%$pts (percentage points) to an average accuracy of $91.7\%$, outperforming the other methods. 

\textbf{Results for Office-Home:} The results for the Office-Home dataset are shown in Tab. \ref{tab:officeHome}. Again, our base network already performs quite well, but with the addition of GSDE, we further increase the accuracy by $2.3\%$pts to an average accuracy of $73.6\%$. We outperform the existing methods, with an increase of almost $1\%$pts over FixBi, the next best-performing algorithm. 

\textbf{Results for DomainNet:} The results for the DomainNet dataset are shown in Tab. \ref{tab:domainNet}. The addition of the GSDE strategy lets us increase the per-class accuracy by almost $2\%$pts to an average per-class accuracy of $83.07\%$. Again we outperform the existing methods, with an increase of $1.68\%$pts over SENTRY, the next best-performing algorithm.

The increase in accuracy with the addition of the GSDE strategy as well as outperforming other domain adaptation methods on all three datasets shows the effectiveness of our proposed method.

\begin{table*}
  \caption{Per class accuracy results on DomainNet dataset. Best results are displayed in bold and the runner-up results are underlined. We display the results of the network with and without our proposed Gradual Source Domain Expansion strategy.}
  \label{tab:domainNet}
  \centering
  \begin{tabular}{llllllllllllll}
    \toprule
    Method & R$\shortrightarrow$C  & R$\shortrightarrow$P & R$\shortrightarrow$S & 
    C$\shortrightarrow$R & C$\shortrightarrow$P  & C$\shortrightarrow$S &
    P$\shortrightarrow$R & P$\shortrightarrow$C  & P$\shortrightarrow$S &
    S$\shortrightarrow$R & S$\shortrightarrow$C  & S$\shortrightarrow$P &
     Avg \\
    \midrule
    ResNet-50 \cite{ResNet} & 58.84 & 67.89 & 53.08 & 76.70 & 53.55 & 53.06 & 84.39 & 55.55 & 60.19 & 74.62 & 54.60 & 57.78 & 62.52 \\
    \midrule
    BBSE \cite{BBSE} & 55.38 & 63.62 & 47.44 & 64.58 & 42.18 & 42.36 & 81.55 & 49.04 & 54.10 & 68.54 & 48.19 & 46.07 & 55.25 \\
	MCD \cite{MCD} & 61.97 & 69.33 & 56.26 & 79.78 & 56.61 & 53.66 & 83.38 & 58.31 & 60.98 & 81.74 & 56.27 & 66.78 & 65.42 \\
	UAN \cite{UAN} & 71.10 & 68.90 & 67.10 & 83.15 & 63.30 & 64.66 & 83.95 & 65.35 & 67.06 & 82.22 & 70.64 & 68.09 & 72.05 \\
	ETN \cite{ETN} & 69.22 & 72.14 & 63.63 & 86.54 & 65.33 & 63.34 & 85.04 & 65.69 & 68.78 & 84.93 & 72.17 & 68.99 & 73.99 \\
	BSP \cite{BSP} & 67.29 & 73.47 & 69.31 & 86.50 & 67.52 & 70.90 & 86.83 & 70.33 & 68.75 & 84.34 & 72.40 & 71.47 & 74.09 \\
	COAL \cite{COAL} & 73.85 & 75.37 & 70.50 & 89.63 & 69.98 & 71.29 & 89.81 & 68.01 & 70.49 & 87.97 & 73.21 & 70.53 & 75.89 \\
	InstaPBM \cite{InstaPBM} & 80.10 & 75.87 & 70.84 & 89.67 & 70.21 & 72.76 & 89.60 & 74.41 & 72.19 & 87.00 & 79.66 & 71.75 & 77.84 \\
	BIWAA-I \cite{BIWA} & 79.93 & 75.24 & 75.35 & 87.93 & 72.07 & 75.71 & 88.87 & 77.81 & 76.66 & 88.78 & 80.49 & \underline{74.49} & 79.44 \\
	Sentry \cite{SENTRY} & \textbf{83.89} & 76.72 & 74.43 & \underline{90.61} & \underline{76.02} & \underline{79.47} & \underline{90.27} & \underline{82.91} & 75.60 & \textbf{90.41} & 82.40 & 73.98 & \underline{81.39} \\
	\midrule
	\textbf{Ours} & 80.72 & \underline{77.96} & \underline{79.71} & 90.19 & 75.61 & 76.01 & 89.26 & 80.74 & \underline{76.97} & 89.27 & \underline{82.65} & 74.47 & 81.13 \\
	\textbf{Ours+GSDE} & \underline{82.93} & \textbf{79.16} & \textbf{80.76} & \textbf{91.92} & \textbf{78.16} & \textbf{79.98} & \textbf{90.92} & \textbf{84.10} & \textbf{79.16} & \underline{90.30} & \textbf{83.36} & \textbf{76.07} & \textbf{83.07} \\
    \bottomrule
  \end{tabular}
\end{table*}

\section{Ablation studies}

\textbf{GSDE with other UDA methods:} \\
\begin{table}
  \caption{Improvements of various UDA methods using GSDE on different datasets. The first row depicts the original accuracy, the second (+) the results with the addition of GSDE, and the third row ($\Delta$) shows the improvement. We report the average accuracy over three seeds using the implementation of the respective publication.}
  \label{tab:gsde}
  \centering
  \begin{tabular}{lllllll}
    \toprule
     & Ours & Sentry & CDAN & +E & AFN & SHOT \\
    \midrule
    \multicolumn{7}{l}{Office-31} \\
    \midrule
    Orig & \textbf{90.65} & 87.26 & 87.51 & \underline{88.80} & 85.35 & 88.06 \\
    + & \textbf{91.65} & 88.73 & 89.53 & \underline{90.17} & 88.06 & 89.19 \\
    $\Delta$ & +1.01 & +1.48 & +2.01 & +1.37 & +2.71 & +1.13 \\
    \midrule
    \multicolumn{7}{l}{Office-Home} \\
    \midrule
    Orig & 71.33 & \textbf{72.11} & 66.50 & 68.65 & 66.67 & \underline{71.99} \\
    + & \underline{73.63} & \textbf{74.06} & 70.85 & 71.90 & 70.54 & 73.24 \\
    $\Delta$ & +2.31 & +1.95 & +4.35 & +3.25 & +3.87 & +1.25 \\
    \midrule
    \multicolumn{7}{l}{DomainNet} \\
    \midrule
    Orig & \underline{81.13} & \textbf{81.64} & 75.79 & 77.01 & 74.81 & 78.81 \\
    + & \textbf{83.07} & \underline{82.54} & 81.16 & 81.62 & 78.61 & 79.14 \\
    $\Delta$ & +1.94 & +0.90 & +5.37 & +4.60 & +3.80 & +0.33 \\
    \bottomrule
  \end{tabular}
\end{table}
We implemented our gradual source domain expansion strategy in various UDA methods. Sentry \cite{SENTRY} is based on self-supervised learning using data augmentations, CDAN(+E) \cite{CDAN} is an adversarial method, AFN \cite{AFN} is based on adapting the feature norm between source and target, and SHOT \cite{SHOT} is a source free domain adaptation method. We chose these methods as they use vastly different adaptation strategies. Our strategy significantly improves the results for all methods. The improvement for SHOT is especially interesting since the training on the source data and adaptation on the target data are done separately (first trained on source data, and then adapted using only target data), supporting our assumption that pre-aligning the classifier with a strong prior from the pseudo-source data helps in the adaptation process.

\textbf{Reinitialization and Source Domain Expansion:} In this part, we investigate the benefits of reinitializing the weights of the network each run. For this, the trained weights are kept from the previous run instead of reinitializing them. We further investigate the benefit of the source domain expansion over simply using the pseudo-labels for a classification loss. While in our proposed method the pseudo-source samples are presented to the adversarial loss $L_{AD}$ and semantic loss $L_{SM}$ as source data, this is not the case in this ablation study - solely a classification loss is added for the subset of target samples that would be added as pseudo-source data.
As can be seen in Tab. \ref{tab:reinit} the reinitializing significantly boosts the performance with a gain of more than $1.5\%$pts for OH and DN over keeping the weights. Using a classification loss over the source expansion gains good improvements, but still the proposed method performs significantly better. This shows the benefits of the pseudo-source data for the adversarial adaptation, helping to guide the domain alignment.
\begin{table}
  \caption{Improvements of reinitialization and source domain expansion. Baseline does not use GSDE. No re-init employs the weights of the previous run. No expansion only employs a classification loss instead of source domain expansion.}
  \label{tab:reinit}
  \centering
  \begin{tabular}{l|lll}
    \toprule
      & O31 & OH & DN \\
    \midrule
 Baseline & 90.65 & 71.29 & 81.13   \\
    \midrule
No re-init & 90.89 & 72.07 &  81.32  \\
    \midrule
No expansion & \underline{91.36} & \underline{72.96}  & \underline{82.84}    \\
    \midrule
Proposed & \textbf{91.65} & \textbf{73.63} & \textbf{83.07} \\    
    \bottomrule
  \end{tabular}
\end{table}

%
%

\textbf{Number of max run $N$:}\\
We evaluated our algorithm against different maximum runs. For this evaluation we excluded the two tasks W$\shortrightarrow$D and D$\shortrightarrow$W since the two domains are very similar (commonly done for this dataset). It can be seen in Fig. \ref{fig:maxRun} that the accuracy steeply increases until around $N=3$. Afterward, the accuracy still increases, but not as steeply, indicating that a high $N$ achieves better results. However, since the computational costs increase linearly with $N$, we chose $N=5$ for our experiments, since it is a good trade-off between gain in accuracy and runtime of the algorithm.

\textbf{Number of bottlenecks $k$:} \\
We evaluated our algorithm for different numbers of bottlenecks $k$. In this experiment, we only trained for a max run of one, and the same subset of Office31 is used as in the previous ablation study. The results can be seen in Fig. \ref{fig:bneck}. We chose a $k=5$ for all other experiments.

\textbf{Accuracy over runs:}\\
The accuracy after each run for $N=5$ for the three datasets is plotted in Fig. \ref{fig:accRun}. It can be seen that the accuracy steadily increases with each run.

 \begin{figure}[t]
   \centering
                \includegraphics[width=0.35\textwidth ]{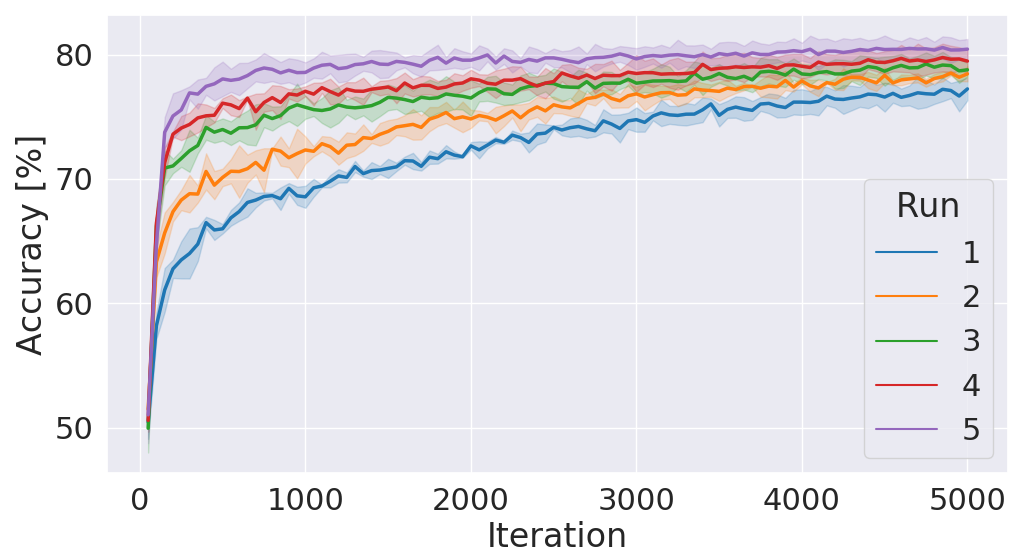}
                \caption{Accuracy for different runs over the training iterations for the task OfficeHome C$\rightarrow$R.}
                \label{fig:iterRun}       
  \end{figure}
  
   \begin{figure}[t]
   \centering

                \includegraphics[width=0.35\textwidth ]{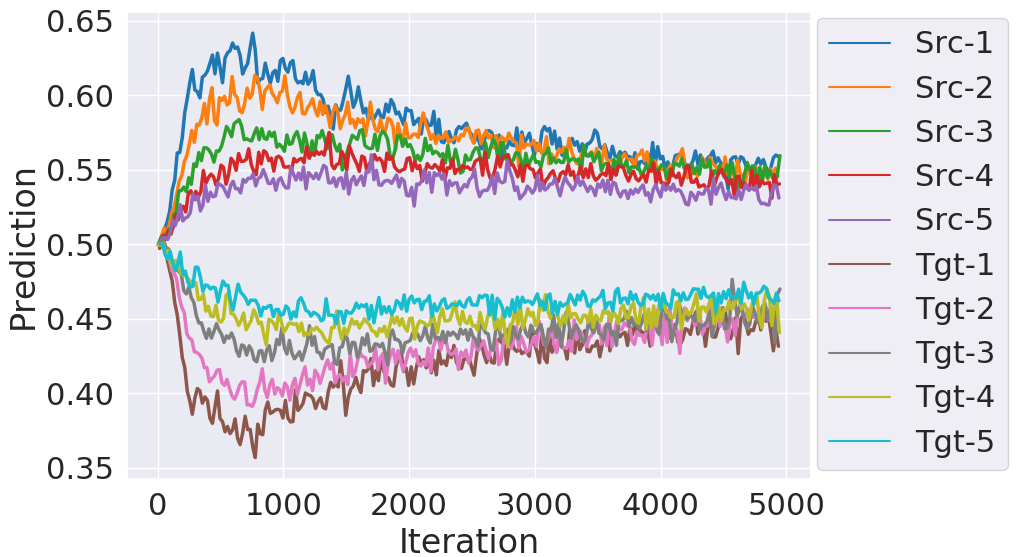}
                \caption{Domain classifier output for different runs over the training iterations. Src represents source data and Tgt represents target data. The number indicates the respective run. The adaptation task was OfficeHome C$\rightarrow$R.}
                \label{fig:domRun}
                \vspace*{-0.5cm}
  \end{figure}

\textbf{Accuracy within runs and domain classifier score:}\\
We plotted the accuracy measured after each $50$ iteration for the adaptation task of C$\shortrightarrow$R in Fig. \ref{fig:iterRun}. It can be seen that the later runs achieve a much higher accuracy early on in the training, showing the effectiveness of the introduced pseudo-source data. This higher accuracy also carries over into the later stages of the training. The averaged output from the domain classifier is plotted in Fig. \ref{fig:domRun}. An output of $1$ represents a discriminator prediction of source domain and $0$ of target domain, respectively. A score of $0.5$ means that source and target are equally likely - the case for domain invariant features. It can be seen that due to the pseudo-source data, the distributions are closer together even early on in the training. 

\textbf{Contribution of each adaptation loss:}
In Tab. \ref{tab:losses} we show the contribution of the three different losses to our base network. For this evaluation, the network is only trained for a max run of one.
\begin{table}
  \caption{Contribution of each adaptation loss.}
  \label{tab:losses}
  \centering
  \begin{tabular}{llll}
    \toprule
    Losses & O31 & OH & DN \\
    \midrule		
    $L_{AD}$ & 87.83 & 65.86 & 76.85 \\
    $L_{AD} + L_{MS}$ & \underline{90.40} & \underline{70.45} & \underline{79.28} \\
    $L_{AD} + L_{MS} + L_{SS}$ & \textbf{90.65} & \textbf{71.33} & \textbf{81.13}\\
    \bottomrule
  \end{tabular}
\end{table}

\textbf{Other improvements:}

In Tab. \ref{tab:gsde} we examine the benefit of the multiple bottleneck (MB) and label scoring (LS) strategy to our algorithm. Using both of the improvements increases the accuracy by 0.53$\%$pts for O31, 0.81$\%$pts for OH, and 0.46$\%$pts for DN. Since only using LS showed a decrease in accuracy for DN, we ran the experiments for DN with 3 additional seeds (total of 6) to decrease the effect of randomness. We believe that the decrease in accuracy can be explained as DN uses the per-class accuracy as reported value (the other datasets use overall accuracy). When changing the evaluation criteria to overall accuracy, there is actually a gain of $0.19\%$pts, indicating that the LS strategy favors high sample classes. However, it is interesting to note that using both strategies achieves a higher gain than adding the gains of each strategy, hinting that there is a good synergy between the two.

\section{Discussion and Limitations}
In this work, we presented a gradual source domain expansion strategy for the unsupervised domain adaptation task. The GSDE strategy introduces a strong prior in the form of pseudo-source data to help align the network early on to the target domain in order to prevent an early alignment build-up error. We show that with our base network, consisting of an adversarial loss, a semantic loss, and a semi-supervised loss, we can increase the performance significantly using the GSDE strategy. We further showed that the GSDE strategy can be applied to a wide range of existing domain adaptation methods significantly increasing the performance.

While the proposed method is effective, one limitation of the algorithm is that the computational costs increase linearly with the amount of runs $N$.

\begin{table}
  \caption{Improvements of multiple bottlenecks (MB) and label scoring (LS) to our method.}
  \label{tab:gsde}
  \centering
  \begin{tabular}{llll}
    \toprule
    Method & O31 & OH & DN \\
    \midrule
    None & 91.12 & 72.83 & 82.81\\
    +LS &  91.21 (+0.09) & 73.00 (+0.17) & 82.69 (-0.12) \\
    +MB & \underline{91.34} (+0.22) & \underline{73.22} (+0.39) & \underline{83.03} (+0.22) \\
    +LS+MB &  \textbf{91.65} (+0.53) & \textbf{73.63} (+0.81) & \textbf{83.27} (+0.46) \\
    \bottomrule
  \end{tabular}
\end{table}

{\small
\bibliographystyle{ieee_fullname}
\bibliography{egbib}
}

\end{document}